\documentclass[10pt,a4paper]{article}
\def\arxivversion{}

\usepackage[utf8]{inputenc}
\usepackage[T1]{fontenc}
\usepackage[american]{babel}
\usepackage{newtxtext,newtxmath}
\usepackage[a4paper,left=2cm,right=2cm,top=2.54cm,bottom=3.76cm]{geometry}
\usepackage{graphicx}
\usepackage{amsmath}
\usepackage{booktabs}
\usepackage{xcolor}
\usepackage{enumitem}
\usepackage{multicol}
\usepackage[hidelinks]{hyperref}
\usepackage{fancyhdr}
\usepackage{titlesec}
\usepackage{caption}

\usepackage[numbers,sort&compress]{natbib}
\graphicspath{{figures/}}

\newcommand{\obpdcfooter}{\ifdefined\arxivversion\else OBPDC 2026\fi}
\fancypagestyle{plain}{\fancyhf{}\fancyfoot[C]{\obpdcfooter}}

\titleformat{\section}{\normalfont\normalsize\bfseries}{}{0pt}{\MakeUppercase}
\titlespacing*{\section}{0pt}{\baselineskip}{\baselineskip}
\titleformat{\subsection}{\normalfont\normalsize\bfseries}{}{0pt}{}
\titlespacing*{\subsection}{0pt}{\baselineskip}{\baselineskip}
\titleformat{\subsubsection}{\normalfont\normalsize\itshape}{}{0pt}{}
\titlespacing*{\subsubsection}{0pt}{0.5\baselineskip}{0.25\baselineskip}

\setlist{nosep,leftmargin=1.2em}

\begin{document}

\begin{center}
{\fontsize{12}{14}\selectfont\bfseries
AI-BASED ON-BOARD MARITIME OBJECT DETECTION FOR EARTH OBSERVATION PAYLOAD DATA REDUCTION ON VERSAL EMBEDDED HARDWARE\ifdefined\arxivversion$^{*}$\fi\par}
\vspace{1.2ex}
{\bfseries Thomas Goudemant, Aur\'elien Bobey, Omar Hlimi, Marjorie Bellizzi\par}
\vspace{1ex}
{\bfseries\itshape IRT Saint Exup\'ery\par}
Email: thomas.goudemant@irt-saintexupery.com
\end{center}
\ifdefined\arxivversion
  {\renewcommand{\thefootnote}{*}%
  \footnotetext{Accepted at the 10th On-Board Payload Data Compression Workshop (OBPDC 2026), Barcelona, Spain, 12--14 October 2026.}}
\fi

\begin{abstract}
Very-high-resolution Earth-observation satellites acquire more data than they can store and downlink, while in maritime surveillance the vessels cover a tiny fraction of each scene.
We study onboard vessel detection as a way to select what is downlinked, which reduces the data according to its content rather than coding every pixel; it is complementary to conventional onboard compression.
The work follows three axes.
(i) Data and algorithm: a controlled dataset is generated from 68 annotated Maxar scenes with 43 vessel classes, and a YOLOX-S detector is trained on it.
(ii) Embedded deployment: the detector is quantized and deployed on the DPU of a Versal VC1902, with a limited loss of detection quality and a processing time of a few seconds per scene.
(iii) Data reduction: we propose several downlink modes, from metadata only (box, class and score of each detection) to image crops around vessels, tiles holding detections, or the whole scene with a degraded background, and estimate from the measured detection errors the trade-off each offers between the vessels kept and the volume downlinked.
On our dense harbor and coastal scenes, tiles keep 98\,\% of the vessels with 29\,\% of the scene volume, and crops 83\,\% with 3\,\%.
\end{abstract}

\section{Introduction}
\label{sec:intro}

Very-high-resolution optical EO satellites acquire scenes of hundreds of megapixels to gigapixels.
Data volumes grow faster than downlink capacity, and the onboard mass memory must buffer the data until a ground station is in view \citep{duggan2025survey}.
In maritime surveillance this pressure is paradoxical: most pixels show open sea, while vessels occupy a very small fraction of the scene.
Downlinking every pixel with the same priority spends most of the capacity on content without mission value; small, sparse vessels on large sea areas are therefore a natural first case for onboard selection.

Conventional onboard compression (CCSDS 122/123 \citep{ccsds122,ccsds123}) reduces the number of bits per pixel but does not know where the useful content is.
An onboard detector adds this knowledge: the satellite can downlink only the detected boxes, image crops around vessels, or the whole scene with a more compressed background (selective coding). Detection suits this purpose: a box locates each object compactly, directly defines the crop or tile to keep and takes a few bytes as metadata, whereas a segmentation mask is costlier to describe and transmit.
Such semantic selection is applied before the existing codecs and complements them; moving processing closer to the sensor in this way reduces latency and downlink volume \citep{zhang2022intelligent}.
Content-dependent reduction already exists in coarser forms: $\Phi$-Sat-1 discards cloudy images on board \citep{giuffrida2022phisat1}, and cloud screening combined with CCSDS-123 avoids transmitting cloudy pixels \citep{cilia2020cloud}. Object detection makes the selection finer, at object level, and specific to the mission target.

Onboard AI has also been demonstrated in orbit, e.g.\ the $\Phi$sat-2 processing chain \citep{melega2023phisat2,104goudemant2024anomaly80531} and OPS-SAT cloud segmentation \citep{feresin2021opssat}.
Onboard object detection has been studied on high-resolution imagery \citep{shen2023onboard}, including YOLOX \citep{ge2021yolox} on an on-orbit FPGA \citep{wang2022yoloxfpga}; ship detection in optical imagery is widely studied on the ground \citep{li2021shipsurvey}. Our team has shown the benefit of onboard YOLO-family detectors \citep{redmon2018yolov3,ge2021yolox} for ships \citep{goudemant2022ships} and for building damage assessment \citep{goudemant2026building}, integrated in a reactive ground--space loop \citep{francesconi2025reactive}.
These works focus on detection quality and embedded feasibility; the data reduction that the detections actually allow, with their errors, on real scenes, is rarely quantified.

This paper quantifies this data reduction, and ties it to the detector performance measured on controlled data and on the target hardware: a YOLOX detector is trained on a controlled maritime dataset generated from annotated Maxar scenes, quantized and deployed on the DPU accelerator of a Versal VC1902 device at one operating point, and its measured errors are used to estimate the downlinked volume of several selection modes on real scenes.

\section{Method}
\label{sec:method}

\begin{figure}[tb]
  \centering
  \includegraphics[width=\linewidth]{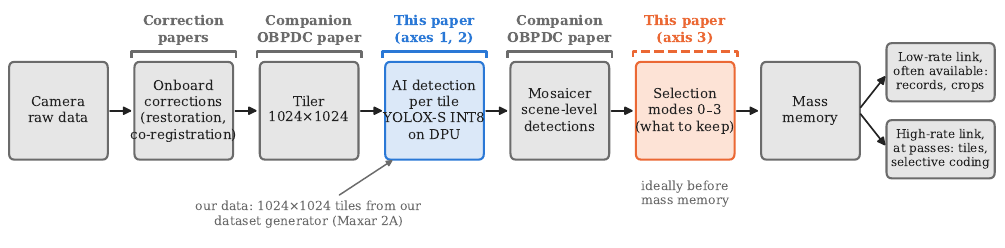}
  \caption{Onboard image flow and scope of this work. Corrections: \citep{dorise2025raw,dorise2026restoration,goudemant2026bitemporal}; tiler and mosaicer: \citep{foulon2026generic}.}
  \label{fig:chain}
\end{figure}

Reducing the downlinked data with AI is not a single algorithm but a processing chain, from the acquisition of an image to its downlink (Fig.~\ref{fig:chain}). The camera delivers raw data, which are first corrected on board, e.g.\ restoration of acquisition effects \citep{dorise2025raw,dorise2026restoration} and co-registration, e.g.\ against a compressed latent reference uplinked from the ground \citep{goudemant2026bitemporal}. The corrected scene, too large to be processed at once, is cut by a tiler into tiles (e.g.\ $1024\times1024$ pixels). Each tile then goes through the AI stage, here vessel detection, which outputs detection products (boxes, classes, scores); a mosaicer gathers the tile results into a scene-level result, and a selection stage uses it to decide what is written to the mass memory and downlinked. Tiling and mosaicking are presented in a companion paper \citep{foulon2026generic}.
Our vessel detection, which enables the data reduction, is thus one stage of this more complete chain; this paper covers the AI stage, which works tile by tile, and the selection it drives.
Our data are already corrected ground products, cut into $1024\times1024$ tiles by our own dataset generator, so neither the onboard corrections nor the tiler are applied here.
Within this scope, the method is organized along three axes: the data and the detector (axis~1), its embedded deployment (axis~2), and the use of its outputs to reduce the downlinked data (axis~3).

\subsection{Axis 1: Data and Algorithm}
\label{sec:dataset}
\begin{figure}[tb]
  \centering
  \includegraphics[width=\linewidth]{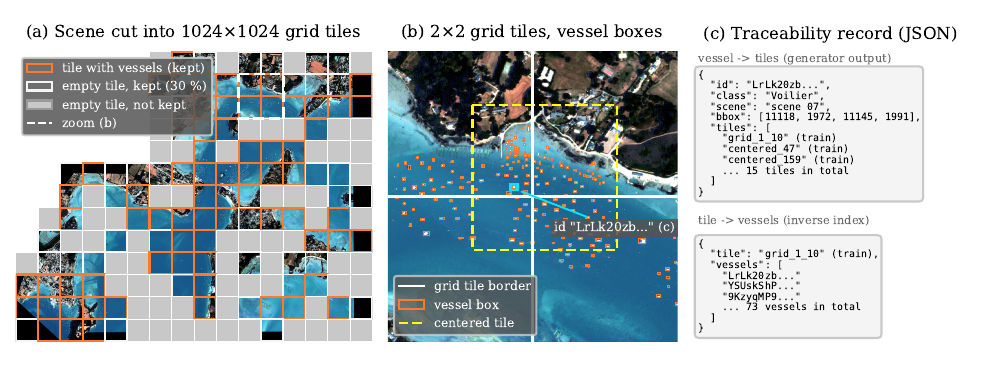}
  \caption{Tiling of one scene: (a) the $1024\times1024$ grid, with the tiles holding vessels, the empty tiles kept (30\,\%) and those not kept; (b) zoom on $2\times2$ grid tiles with the vessel boxes and one vessel-centered tile; (c) traceability record of the highlighted vessel, which appears in one grid tile and several centered tiles, and the inverse index listing the vessels of that grid tile. Images shown at 1/8 and 1/2 of the native resolution. Maxar Products Standard (2A) Imagery \textcopyright{} 2017--2024 Maxar Technologies.}
  \label{fig:tiling}
\end{figure}

\emph{Controlled dataset generation.} Cutting annotated scenes naively gives a dataset dominated by empty background, poor in rare classes, with neighboring tiles shared between training and test, and with no link back to the scenes. We therefore developed a controlled generation algorithm that turns annotated scenes into training and evaluation tiles, so that these choices are explicit and measurable and the scores can be trusted and traced back to the scenes. Its input is 68 Maxar Standard (2A) pansharpened scenes (4 bands, 16-bit; RGB used) of harbors, coastal areas with dense traffic and canals, with a ground sampling distance of 30 to 70~cm, annotated by an expert with 43 vessel classes (42 types and a generic ``vessel'' class).
Each scene is converted to 8 bits with a per-scene 2--98\,\% percentile stretch and cut on a regular $1024\times1024$ grid, the same grid as the onboard tiler (\emph{grid tiles}, Fig.~\ref{fig:tiling}b).
The generator then controls four properties of the tiles it produces.

\begin{itemize}
  \item \emph{Empty tiles.} Every grid tile containing a vessel is kept, but only 30\,\% of the empty tiles, drawn at random, so that the background does not dominate (Fig.~\ref{fig:tiling}a).
  \item \emph{Class balance.} Extra tiles centered on vessels (\emph{centered tiles}) are added: on a few vessels of each dense tile (about $\sqrt{N}/2$ for $N$ vessels), and on vessels cut by the grid. This adds views of rare classes and of whole vessels without adding one tile per vessel, which would over-represent moored ships.
  \item \emph{Spatial split.} Blocks of neighboring tiles are assigned to training, validation or test (80/10/10\,\%), so that a vessel normally belongs to a single subset. The centered tiles can however overlap a block of another subset, so some vessels appear in two subsets.
  \item \emph{Traceability.} Every vessel keeps an identifier linking it to all tiles that contain it, to its scene and to its scene coordinates; inverting this record gives the vessels of each tile (Fig.~\ref{fig:tiling}c).
\end{itemize}

\emph{Detector training and evaluation.} On these tiles we train YOLOX-S \citep{ge2021yolox}, a recent member of the YOLO family \citep{redmon2015yolo} chosen for its accuracy/cost trade-off and its compatibility with the Vitis AI flow; it predicts a box, a class and a confidence for each vessel, and is trained in floating point for 100 epochs. The validation tiles later serve to calibrate the quantization (axis~2), and the test tiles, restricted to grid tiles as produced by the onboard tiler, only to measure the scores. Since only finding a vessel matters for data retention, detection is evaluated separately from classification, with the usual IoU $\geq 0.5$ matching, as a function of the confidence threshold. Scores are reported at the deployed threshold of 0.30, which balances missed vessels and false alarms; the data reduction (axis~3) uses a lower one, which favors keeping vessels.

\subsection{Axis 2: Embedded Deployment}
\label{sec:detector}
\emph{Target and quantized model.} The detector is trained in floating point, but the target computes with 8-bit integers (int8): the model must therefore be quantized before it is deployed, which affects its detection performance. The target is the VCK190 evaluation board, built around a Versal VC1902 device \citep{gaide2019versal}, an adaptive-compute architecture evaluated for space edge computing \citep{perryman2023versal}. It combines Arm Cortex-A72 cores with a DPU (deep-learning processing unit) accelerator on the AI Engines and programmable logic. The model is quantized by Vitis AI post-training quantization, calibrated on the validation tiles, then compiled for one operating point: DPUCVDX8G in its C32B6 configuration (six batch handlers of 32 AI Engine cores, one compute unit), i.e.\ batches of six tiles, with a $1024\times1024$ input (Vitis AI~3.0). The effect of quantization and deployment is measured by comparing the quantized model with the float one, on the test set and with the same metrics.

\emph{Inference pipeline and hardware measurements.} The C++ inference code runs on the Arm cores and drives the DPU through the Vitis AI runtime, in three stages per batch: pre-processing (CPU) scales the 8-bit tiles to the quantized model input (input scale); inference runs on the DPU; post-processing (CPU) does what the network does not, i.e.\ output rescaling, box decoding, confidence threshold and per-class non-maximum suppression (NMS). All processing stays at tile level, and reading the tiles from the board storage (an SD card) is not counted, since on board they would arrive as a stream from the tiler. Over these three stages, the latency of each stage is measured per batch, the throughput is derived in Mpixel/s, and the board power is read from its INA226 power-monitoring chips, sampled every 2~ms. Together they give the energy needed per tile and per scene, split into a static part (board at rest) and a dynamic part (increase during processing).

\subsection{Axis 3: Data Reduction by Onboard Filtering}
\label{sec:reduction}
\begin{figure}[tb]
  \centering
  \includegraphics[width=\linewidth]{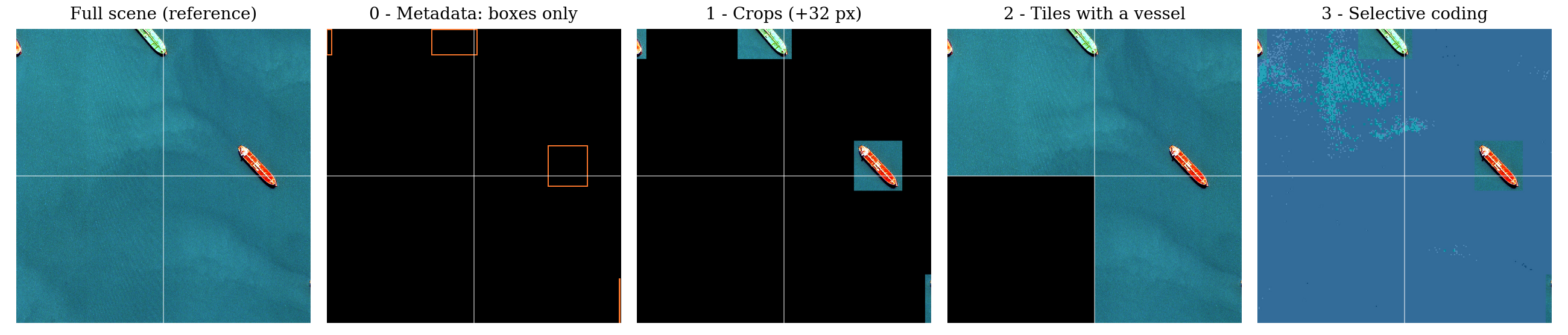}
  \caption{Downlinked content of each mode on $2\times2$ grid tiles: full scene (reference); metadata, i.e.\ detected boxes only; crops with a 32-pixel margin; grid tiles holding a detection; selective coding, with the crops at full quality and the background compressed. The ship-tile examples use Maxar Products Standard (2A) Imagery \textcopyright{} 2017--2024 Maxar Technologies.}
  \label{fig:modes}
\end{figure}

\emph{Downlink modes and data model.} The detections of the deployed model then decide what is downlinked. Four modes are compared, from the most to the least reduced (Table~\ref{tab:modes}, Fig.~\ref{fig:modes}). \emph{Metadata} sends only one record per detection (box, class, score), with no image: the ground learns where the vessels are, but cannot see them. \emph{Crops} adds a small image around each detected box, the cheapest way to see the vessels. \emph{Tiles} sends every grid tile holding at least one detection, which also gives the vessels' surroundings and tolerates a missed vessel next to a detected one. \emph{Selective coding} sends the whole scene, with the crops at full quality and the background compressed with a ratio CR, so that nothing is discarded. They are compared with the full scene, which is the usual product, before any onboard compression. For each mode, the table gives the downlinked content, its size in bytes, the main risk due to detection errors, and what sets its volume. The modes can also be combined in time over downlinks of different rates and availability, e.g.\ a low-rate link available often and a high-rate link during ground-station passes \citep{kodheli2021satcom} (Fig.~\ref{fig:chain}): the most reduced products give the ground an early view of the scene, and less reduced ones follow, possibly on ground request. We quantify the volume of each product, not the link schedule.

\begin{table}[tb]
  \centering\small
  \caption{Downlink modes compared in this work. $m$: crop margin; CR: background compression ratio. Size: images as raw data of 12 bits per band, packed (4.5 bytes per pixel, 3 bands); a detection record holds two box corners in tile coordinates ($4\times10$ bits), the class (6 bits), the score (8 bits) and the tile index (11 bits), i.e.\ 65 bits.}
  \label{tab:modes}
  \setlength{\tabcolsep}{4pt}
  \begin{tabular}{lllll}
    \toprule
    Mode & Downlinked content & Size & Main risk & Volume set by \\
    \midrule
    0 -- metadata & one record per detection & 9 B per record & no image & number of detections \\
    1 -- crops & box + margin $m$ = 32 px & 4.5 B per pixel & missed vessels lost & vessel area, $m$, false alarms \\
    2 -- tiles & grid tiles with $\geq$ 1 detection & 4.5 B per pixel & missed tiles lost & kept tiles \\
    3 -- selective coding & crops + background at CR & crops + rest\,/\,CR & missed vessels compressed & crop area, CR \\
    reference & full scene & 4.5 B per pixel & none & scene size \\
    \bottomrule
  \end{tabular}
\end{table}

\emph{From detection errors to downlinked volume.} The detector makes two kinds of errors: false negatives (missed vessels), which may lose information, and false positives (false alarms), which only add volume. Their effect depends on the mode. A missed vessel is lost with metadata and crops, survives in a tile if a neighbor in the same tile is detected, and remains, compressed, in the background with selective coding. A false alarm costs a record or a crop, but a tile only when it falls on an empty tile. Since a lost vessel cannot be recovered on the ground, whereas a false alarm only costs volume, the estimation uses a lower threshold than the deployed one (0.10 instead of 0.30). This is where the previous steps come in: the errors of the deployed detector, measured on the test set, are applied to the tiles and vessels of each of the 68 scenes to estimate the order of magnitude of the downlinked volume and of the vessels lost; for selective coding, the background compression ratio CR is treated as a parameter, from mild to strong compression. The result depends strongly on the scenes. Ours are harbors, coastal areas with dense traffic and canals, which is a nominal case when such areas are tasked; on open sea, vessels are much sparser and the reduction would be much larger, so our figures are conservative.

\section{Experiments and Results}
\label{sec:experiments}

\subsection{Axis 1: Data and Algorithm}
\label{sec:res-data}
\begin{figure}[tb]
  \centering
  \begin{minipage}[b]{0.27\linewidth}
    \centering\footnotesize
    \captionof{table}{Dataset split (68 scenes). A vessel in two subsets is counted in both. Test: grid tiles only.}
    \label{tab:dataset}
    \setlength{\tabcolsep}{3pt}
    \begin{tabular}{lrrr}
      \toprule
      & Train & Val. & Test \\
      \midrule
      Tiles & 11\,954 & 1\,119 & 894 \\
      Empty & 4\,719 & 437 & 470 \\
      Vessels & 21\,471 & 4\,146 & 1\,596 \\
      Classes & 43 & 32 & 33 \\
      \bottomrule
    \end{tabular}
    \vspace{6pt}
  \end{minipage}\hfill
  \begin{minipage}[b]{0.71\linewidth}
    \centering
    \includegraphics[width=\linewidth]{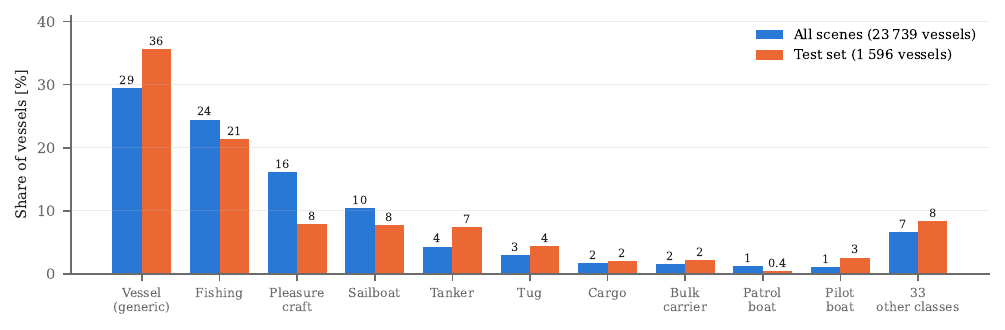}
    \caption{Share of vessels per class, over the 68 scenes and in the test set. The ten classes shown are those used for the reported mAP.}
    \label{fig:classes}
  \end{minipage}
\end{figure}

\emph{Generated dataset.} The generator produces 13\,967 tiles (Table~\ref{tab:dataset}); keeping 30\,\% of the empty grid tiles leaves two fifths of empty tiles in the dataset, enough to learn the background without letting it dominate. Even after class balancing, the class distribution stays very unbalanced (Fig.~\ref{fig:classes}): four classes hold 80\,\% of the vessels, and the 33 least frequent classes each less than 1\,\%. The rarest classes occur in only a few scenes, so the spatial split cannot place all of them in every subset; the test set nevertheless roughly follows the overall distribution. It holds 1\,596 vessels; since each tile is processed on its own, the 53 vessels cut by the grid are scored in each of their tiles, i.e.\ 1\,649 boxes.

\begin{figure}[tb]
  \centering
  \includegraphics[width=0.7\linewidth]{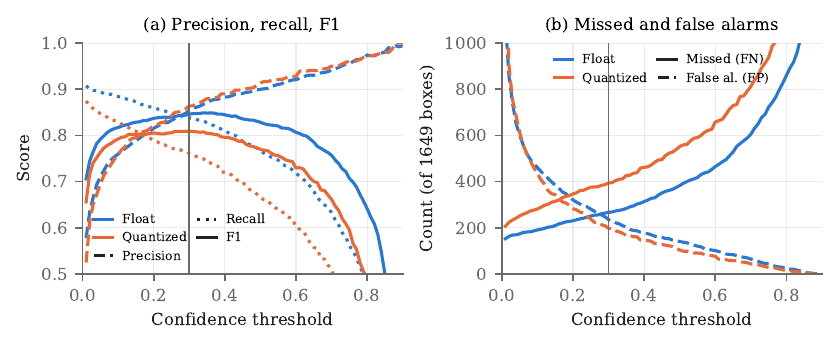}
  \caption{Detection on the test set versus confidence threshold, float and quantized models: (a) precision, recall and F1; (b) missed vessels and false alarms. Vertical line: deployed threshold 0.30.}
  \label{fig:pr}
\end{figure}

\emph{Detection performance of the float model.} Before deployment, the detector finds vessels reliably: on the test set it reaches an AP of 86\,\% and an F1 of 85\,\% at the deployed threshold, with similar precision and recall (Table~\ref{tab:algo}). Naming the vessel type is much harder: one detected vessel in seven gets a wrong type. Classification is also hard to measure on rare classes, so the mAP is averaged over the ten most frequent ones (91\,\% of the test vessels). Since only detection drives data reduction, the rest of the paper uses detection scores. Their dependence on the threshold is weak around the deployed value: the F1 is maximal near 0.30 and changes by only 2 to 4 points between 0.2 and 0.5 (Fig.~\ref{fig:pr}a), the threshold mainly trading missed vessels against false alarms (Fig.~\ref{fig:pr}b). Choosing it is therefore less a matter of accuracy than of which error the mission prefers, which the data reduction exploits.

\begin{table}[tb]
  \centering\footnotesize
  \begin{minipage}[t]{0.36\linewidth}
    \centering
    \caption{Test set (894 tiles, 1\,596 vessels), IoU $\geq$ 0.5, scores in \%. AP and mAP do not depend on the threshold; other scores at 0.30. Class error: share of detected vessels with a wrong type.}
    \label{tab:algo}
    \setlength{\tabcolsep}{3pt}
    \begin{tabular}{lrrr}
      \toprule
      & Float & Quant. & $\Delta$ \\
      \midrule
      \multicolumn{4}{l}{\emph{Detection (class ignored)}} \\
      AP & 86.1 & 81.4 & $-$4.7 \\
      Precision & 85.4 & 86.4 & +1.0 \\
      Recall & 83.8 & 76.1 & $-$7.7 \\
      F1 & 84.6 & 80.9 & $-$3.7 \\
      Missed [count] & 267 & 394 & +127 \\
      False alarms [count] & 236 & 198 & $-$38 \\
      \midrule
      \multicolumn{4}{l}{\emph{Classification (type)}} \\
      mAP, 10 classes & 68.9 & 59.0 & $-$9.9 \\
      mAP, all classes & 50.6 & 43.4 & $-$7.2 \\
      Class error & 14.1 & 14.4 & +0.3 \\
      \bottomrule
    \end{tabular}
  \end{minipage}\hfill
  \begin{minipage}[t]{0.6\linewidth}
    \centering
    \caption{Hardware performance at the selected operating point (VCK190, batches of six $1024\times1024$ tiles). Static: board at rest; dynamic: increase during processing.}
    \label{tab:hw}
    \setlength{\tabcolsep}{3pt}
    \begin{tabular}{lrrr}
      \toprule
      & \multicolumn{2}{c}{Time [ms]} & Throughput \\
      \cmidrule(lr){2-3}
      Stage & per batch & per tile & [Mpix/s] \\
      \midrule
      Pre-processing (CPU) & 51.18 & 8.53 & 123 \\
      DPU inference & 50.65 & 8.44 & 124 \\
      Post-processing (CPU) & 0.94 & 0.16 & 6\,690 \\
      Total & 102.77 & 17.13 & 61 \\
      \midrule
      Power, static + dynamic [W] & \multicolumn{3}{r}{26.6 + 3.8 = 30.4} \\
      Energy per tile, static + dynamic [J] & \multicolumn{3}{r}{0.46 + 0.06 = 0.52} \\
      Time per average scene (327 tiles) [s] & \multicolumn{3}{r}{5.6} \\
      Energy per average scene, static + dynamic [J] & \multicolumn{3}{r}{149 + 21 = 170} \\
      \bottomrule
    \end{tabular}
  \end{minipage}
\end{table}

\subsection{Axis 2: Embedded Deployment}
\label{sec:res-embedded}
\emph{Detection performance after deployment, compared with the float model.} Once quantized and deployed, the detector keeps most of its performance: its AP drops from 86 to 81\,\% and its F1 from 85 to 81\,\% (Table~\ref{tab:algo}). Quantization does not make it less precise, but more cautious: precision is unchanged or slightly higher, while recall drops by about 8 points. The quantized model gives lower confidence scores to the vessels it finds (three quarters of the vessels found by both models get a lower score), so at a given threshold more of them fall below it, and the recall is lower at every threshold (Fig.~\ref{fig:pr}). The loss is therefore largely a shift of the operating point rather than a loss of information: lowering the threshold to 0.10 recovers most of the missed vessels, at the cost of more false alarms, which is the trade-off used for the data reduction. Classification suffers more, for two reasons. Non-maximum suppression is applied per class, so when the model hesitates between two types, two boxes of different classes can remain on the same vessel; the quantized model does this more often, and the extra box counts as a false alarm for its class. In addition, the mAP gives the same weight to every class, so a few rare classes weigh heavily in the loss. For data reduction, which only uses detection, the effect of deployment thus amounts to a recall loss that the threshold can largely compensate.

\emph{Hardware performance.} The chain processes a tile in about 17~ms, i.e.\ an average scene in a few seconds (Table~\ref{tab:hw}). The time is split evenly between CPU pre-processing and DPU inference, while post-processing is negligible: the accelerator is not the bottleneck. The pre-processing already uses a look-up table and SIMD instructions, but runs on a single Arm core: spreading it over several cores or overlapping it with inference would nearly double the throughput. The energy is dominated by the static consumption: processing adds only about a seventh to the power at rest, so the energy per scene mostly depends on how long the board is powered. Although measured on the development kit, this static power is representative: that of the final board differs by only about 2\,\%. Reading the tiles from the board storage, excluded here, is slower than processing them, which confirms that the input data flow must come from the onboard chain rather than from storage.

\subsection{Axis 3: Data Reduction}
\label{sec:res-reduction}
\emph{Missed vessels and false alarms, extrapolated to the scenes.} The reduction is evaluated on the 68 whole scenes (Table~\ref{tab:scenes}). With a perfect detector, the tile mode would keep about a quarter of the scene and the crop mode a few percent; both bounds vary by one to two orders of magnitude between scenes (range in Table~\ref{tab:scenes}), so the gain depends first on how dense the scene is. Ours are dense, and open-sea acquisitions would give much larger reductions. The real detector moves away from these bounds through its missed vessels and false alarms. Their rates are measured on the test tiles with the quantized detector, as in Table~\ref{tab:algo} and Fig.~\ref{fig:pr} but at threshold 0.10, then extrapolated to the scenes by applying them to the tile and vessel counts of each scene; they act very differently in the two image modes (Table~\ref{tab:errors}). With crops, every missed vessel is lost: 17\,\% of the vessels at threshold 0.10. With tiles, a missed vessel is lost only if no other vessel of its tile is detected; in our dense scenes this is rare, so the 7\,\% of tiles lost hold only 2\,\% of the vessels. The vessels still lost are isolated ones: on open sea, where a tile usually holds a single vessel, tiles would lose about as many vessels as crops. False alarms, on the other hand, add volume in both modes: a fifth of the downlinked tiles and over a quarter of the crop volume hold only false alarms, less at a higher threshold.

\begin{table}[tb]
  \centering\footnotesize
  \begin{minipage}[t]{0.40\linewidth}
    \centering
    \caption{Scenes used for the data-reduction analysis (68 scenes). Size: pixels holding image data. Raw data: 4.5 bytes per pixel (12 bits, 3 bands). Crop area: union of the vessel boxes with a 32-pixel margin.}
    \label{tab:scenes}
    \setlength{\tabcolsep}{2pt}
    \begin{tabular}{lrcr}
      \toprule
      & Average & Range & Sum over \\
      & scene & over scenes & 68 scenes \\
      \midrule
      Size [Mpix] & 290 & 36--1\,023 & 19\,700 \\
      Raw data [GB] & 1.30 & 0.16--4.6 & 89 \\
      Vessels & 349 & 9--1\,539 & 23\,739 \\
      \midrule
      \multicolumn{4}{l}{\emph{Volume with a perfect detector}} \\
      \textbf{Tiles with a vessel [\%]} & \textbf{24.3} & 5.9--60 & \\
      \textbf{Crop area [\%]} & \textbf{2.4} & 0.07--7.4 & \\
      \bottomrule
    \end{tabular}
  \end{minipage}\hfill
  \begin{minipage}[t]{0.56\linewidth}
    \centering
    \caption{Statistics of missed vessels and false alarms in the two image modes, quantized detector, at the threshold used for data reduction (0.10) and at a stricter one (0.30). Missed vessels: measured on the test tiles. False alarms: rates measured on the test tiles, extrapolated to the 68 scenes.}
    \label{tab:errors}
    \setlength{\tabcolsep}{4pt}
    \begin{tabular}{lcc@{\hspace{8pt}}cc}
      \toprule
      & \multicolumn{2}{c}{\textbf{Threshold 0.10}} & \multicolumn{2}{c}{Threshold 0.30} \\
      \cmidrule(lr){2-3}\cmidrule(lr){4-5}
      & crops & tiles & crops & tiles \\
      \midrule
      \emph{Missed vessels} & & & & \\
      vessels lost [\% of vessels] & \textbf{17} & \textbf{2} & 24 & 5 \\
      tiles discarded [\% of tiles with vessels] & -- & \textbf{7} & -- & 15 \\
      \midrule
      \emph{False alarms} & & & & \\
      downlinked volume wasted [\%] & \textbf{28} & \textbf{21} & 17 & 12 \\
      \bottomrule
    \end{tabular}
  \end{minipage}
\end{table}

\begin{figure}[tb]
  \centering
  \includegraphics[width=0.72\linewidth]{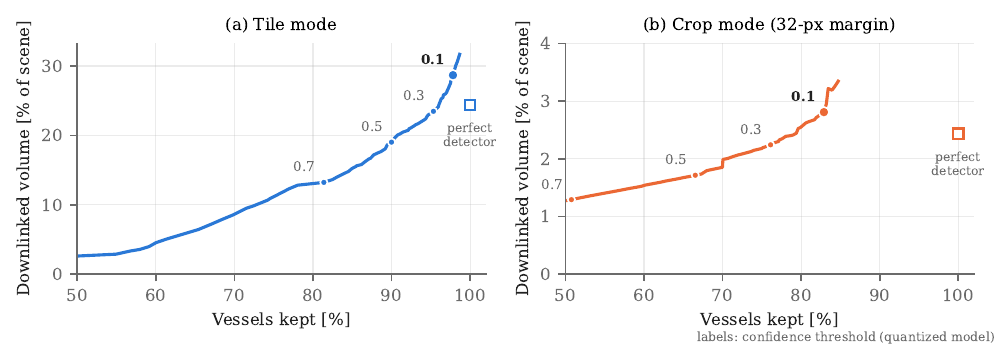}
  \caption{Downlinked volume versus vessels kept, quantized detector, confidence threshold swept (labels; large dot: 0.10, used in Table~\ref{tab:reduction}): (a) tile mode, (b) crop mode. Squares: perfect detector.}
  \label{fig:tradeoff}
\end{figure}

\emph{Volume and vessels kept per mode.} Put together, these errors give, for each mode, the volume downlinked for an average scene and the share of vessels kept (Table~\ref{tab:reduction}), i.e.\ the trade-off between reduction and recall that each mode offers. Metadata take a few kilobytes but show no image; crops give an image of four fifths of the vessels for 3\,\% of the volume, but lose every missed vessel; tiles keep nearly all vessels for under a third of the volume; selective coding loses none, missed vessels remaining in the compressed background, at a volume set by the assumed compression ratio. At this low threshold, false alarms account for 21 to 28\,\% of the volume of the first three modes. The threshold moves each mode along its trade-off curve (Fig.~\ref{fig:tradeoff}): raising it to 0.30 saves about a fifth of the volume of tiles or crops, but increases the vessels lost, much more for crops than for tiles. The background compression ratios are chosen as representative operating points rather than measured with a codec: lossless CCSDS~122.0 coding needs about 4 to 5 bits per pixel on typical EO test images \citep{ccsds120green}, i.e.\ a ratio of about 2.5 to 3 for 12-bit data, so CR~=~4 stands for a mildly lossy background and CR~=~20 for a strongly compressed one.

\begin{table}[tb]
  \centering\small
  \caption{Downlinked data per mode for an average scene (1.30~GB of raw data), quantized detector at \textbf{threshold 0.10}, crops with a 32-pixel margin. Selective coding: volume = crop share $+$ (1 $-$ crop share)/CR, the background being the scene outside the crops; CR is an assumed compression ratio relative to the 12-bit raw data. Volume of false alarms: with metadata, every detection costs 9 bytes, so this is the share of false detections (25\,\%). With crops it is slightly higher (28\,\%): in dense areas, the crops of neighboring vessels overlap and share their 32-pixel margins, whereas each false-alarm crop is isolated and keeps its full margin, so a false alarm costs on average more pixels than a detected vessel.}
  \label{tab:reduction}
  \setlength{\tabcolsep}{5pt}
  \begin{tabular}{lrrrcccc}
    \toprule
    & & & & \multicolumn{4}{c}{Vessels [\% of all vessels]} \\
    \cmidrule(lr){5-8}
    & & & Volume of & seen in & only in & box & \\
    & Data & Volume & false alarms & full & compressed & only & \\
    Mode & [MB] & [\% of full scene] & only [\%] & quality & background & (no image) & lost \\
    \midrule
    0 -- metadata & 0.004 & 0.0003 & 25 & -- & -- & 83 & 17 \\
    1 -- crops & 37 & 2.8 & 28 & 83 & -- & -- & 17 \\
    2 -- tiles & 374 & 28.7 & 21 & 98 & -- & -- & 2 \\
    3 -- selective coding, CR = 4 & 353 & 27.1 & -- & 83 & 17 & -- & 0 \\
    3 -- selective coding, CR = 20 & 100 & 7.7 & -- & 83 & 17 & -- & 0 \\
    reference: full scene & 1\,304 & 100 & -- & 100 & -- & -- & 0 \\
    \bottomrule
  \end{tabular}
\end{table}

\section{Discussion}
\label{sec:discussion}

\emph{Results.} The mode and the threshold, set by the mission, decide how much is sent and how many vessels may be lost. The gain is paid mainly in energy, to be weighed against the transmission energy and storage avoided, which depend on how often the satellite acquires and downlinks. As this energy is mostly static, processing faster and powering the processor down between acquisitions, if possible, saves more than lowering dynamic power. The gain is largest if detection runs before storage, which our tile-by-tile pipeline allows.

\emph{Limitations.} The study covers a single application and a single DPU configuration; since static power dominates, the DPU size should trade its power cost against the throughput it brings. It uses corrected ground products rather than raw data. The scene-level figures extrapolate error rates measured on test tiles, selective coding uses representative compression ratios rather than a codec, and our dense scenes do not represent open sea. Downlink orchestration is not studied.

\emph{Perspectives.} The main next step is raw data, with detection integrated among the other bricks of the onboard chain (Fig.~\ref{fig:chain}): restoration \citep{dorise2025raw,dorise2026restoration}, co-registration, e.g.\ with a compressed latent reference \citep{goudemant2026bitemporal}, tiler and mosaicer \citep{foulon2026generic}, and an onboard codec, with open-sea scenes. The modes could be orchestrated over links of different rates, metadata first; the threshold could adapt to the scene density, and the vessel class set the priority (e.g.\ warships first) once classification is more reliable. Larger networks (e.g.\ YOLOX-L, at the cost of throughput), quantization-aware training and land or cloud masking could improve detection. Selecting only vessels also discards content other users may need, e.g.\ damaged buildings \citep{goudemant2026building}. Open-vocabulary detectors, steered by a text query, could make the selection follow the user's request rather than a fixed class list; they exist in real time \citep{cheng2024yoloworld} and for remote sensing \citep{pan2025laedino}.

\section{Conclusion}
\label{sec:conclusion}
We studied onboard vessel detection to choose what a satellite downlinks. We built a controlled dataset from Maxar scenes, trained YOLOX-S on it and deployed it, quantized, on a Versal DPU. Deployment preserves most detection performance but shifts the operating point toward lower confidence and recall, which a lower threshold largely compensates; an average scene is processed in a few seconds. The detections support several downlink modes, trading data volume against vessels kept: on our dense harbor and coastal scenes, sending crops divides the volume by about 35 while keeping four vessels in five, and sending tiles divides it by about 3.5 while keeping almost all of them. Our main contribution is to tie these steps together: onboard reduction must be judged from the errors of the deployed model and how each mode absorbs them, not from detection scores alone. Raw data in the complete onboard chain, open sea and other applications are the next steps.

\section{References}
\renewcommand{\bibsection}{}
\raggedcolumns
\begin{multicols}{2}
\scriptsize
\setlength{\bibsep}{0pt}
\bibliography{references}
\end{multicols}

\end{document}